# Feature extraction and classification algorithm, which one is more essential? An experimental study on a specific task of vibration signal diagnosis

Qiang Liu[1] and Jiade Zhang[2] and Jingna Liu[3*] and Zhi Yang[1]

[1] Department of Electronic and Communication Engineering, North China Electric Power University, Baoding 071003, Hebei, China
[2] College of Mathematics and Information Science, Hebei University, Baoding 071002, Hebei, China
[3*] State Key Laboratory of Mechanical Behavior and System Safety of Traffic Engineering Structures, Shijiazhuang Tiedao University, Shijiazhuang 050043, China
`*Corresponding author. E-mail: 1263946309@qq.com`

**Abstract.** With the development of machine learning, a data-driven model has been widely used in vibration signal fault diagnosis. Most data-driven machine learning algorithms are built based on well-designed features, but feature extraction is usually required to be completed in advance. In the deep learning era, feature extraction and classifier learning are conducted simultaneously, which will lead to an end-to-end learning system. This paper explores which one of the two key factors, i.e., feature extraction and classification algorithm, is more essential for a specific task of vibration signal diagnosis during a learning system is generated. Feature extractions from vibration signal based on both well-known Gaussian model and statistical characteristics are discussed, respectively. And several classification algorithms are selected to experimentally validate the comparative impact of both feature extraction and classification algorithm on prediction performance.

## 1 Introduction

The gearbox, as one of the most important basic components of an equipment set, has been widely used in automobiles, aviation, machinery industry, wind turbine, railway, and so on [1]. Since the use of gearboxes is often with complex processing technology, high assembly accuracy, and different operating environments, the comments of the gearbox are usually easy-damaged [2]. The gearbox failure will cause the whole equipment to fail to operate normally, and the consequences are dire. It implies that the gearbox fault diagnosis is a really practical but challenging topic.

The research on gearbox fault diagnosis can trace back to the 1960s. Many methodologies and technologies about vibration signal processing can be found from refer-ences [3,4,5]. Interestingly, up to today, this topic is still attracting more scholars and engineers. Actually, the vibration signal of rotating parts, which generally reflects the health status of machinery, is mainly measured by wave equipment and recorded as a wave in the field of gearbox fault diagnosis.

The earlier research on gearbox failure detection focused on the use of spectrum, amplitude, and phase modulation techniques to detect different types of gearbox faults, which was mainly realized through time-domain or frequency-domain analysis methods [6, 7]. However, the actual working environment of the gearbox may be very complex, which is often accompanied by severe noise interference, and existing analysis methods may not perform well for its fault diagnosis. To solve this problem, some improved methods based on time-frequency analysis (e.g., Short-time Fourier analysis, Wigner Ville distribution, and Wavelet analysis, etc.) are proposed [8, 9], which expand the scope of fault detection and possess the ability of noise-resisting to some extent.

Additionally, different models have been built to eliminate the limitation of the uncertainty principle in time-frequency analysis. For instance, Huang et al. in [10] proposed the Empirical Mode Decomposition (EMD) method. Furthermore, Wang et al. in [11] developed an approach to combining ensemble local mean decomposition (ELMD) and singular value decomposition (SVD) to diagnose the early fault of the gearbox. It basically solved the problem of mode aliasing in the EMD method.

With the development of machine learning techniques, many researchers and engineers have begun to consider gearbox fault diagnosis as a classification problem. Based on data-driven models, most machine learning methods can adaptively capture the essential properties of data and the implicit connections between data, and thus, can significantly improve the efficiency of fault diagnosis. For the data-driven model, transforming the original signal into digital features has become a critical step of fault diagnosis. This is the phase of feature extraction, which is usually completed before performing a classification algorithm in traditional fault diagnosis. In the contemporary era of deep learning, feature extraction is still before performing learning algorithms but is iterative and dynamic, which may be the main reason of deep learning models' high accuracy. [12].

It is well acknowledged that feature extraction is one of the most crucial steps in signal fault diagnosis. Many methodologies for signal feature extraction can be found from references. For example, Davis et al. ([13]) in 1980 proposed Mel Frequency Cepstral Coefficient (MFCC), which is widely used in speech recognition for signal digital feature extraction. Mahaphonchaikul et al. ([14]) used Wavelet Transform (WT) to process the EMG signal and then extracted the root mean square, the logarithm of root mean square, the centroid of frequency, and standard deviation as digital features of the signal. Awang et al. ([15]) developed an eigenvector method to obtain the maximum, minimum, mean, and standard deviation of power spectral density, which are considered as the EEG signal's digital features. Chen et al. ([16]) and Lu et al. ([17]) used two-term and three-term Gaussian models to extract features of pulse wave signals, respectively. Based on the types of extracted features, the current digital feature extraction methods can be roughly categorized into two classes. One type is statistical, while the other is of curve fitting. For example, the statistical values are computed after signal preprocessing in [13, 14, 15]; and the curve of function is used in [16, 17] to fit and reconstruct the signal waveform and then to take the parameters of the reconstructed function as the signal's digital features.

Further studies on this topic have confirmed that the methods mentioned above of feature extraction have some limitations [18]. It is criticized that some methods based on signal feature preprocessing have the poor ability of anti-interference. For example, the MFCC proposed in [19] is proved to be not robust to speech signals in noisy environments. The feature extraction based on wavelet is easily affected by the adjacent harmonic components in the signal, which may directly downgrade the fault diagnosis' performance [20]. Due to the limitation of fitted curve shape, the methods based on the curve fitting possibly bring more additional errors [21]. In order to improve the stability of data and the generalization performance of unseen data, Wu et al. ([22]) discussed a feature reorganization scheme based on CCA feature fusion and normalization extraction, and they experimentally verified the effectiveness of the scheme. Li et al. ([23]) used the stacking method regarding ten feature clusters as a view applied in the same predictor or classifier to produce one stacking feature.

In view of the poor anti-interference performance of signal statistical characteristics and the limitations of curve fitting methods, we try to develop an approach to jointly use both types of features extracted from vibration signals given a task of gearbox fault diagnosis. After feature extraction, the signal fault diagnosis will be transformed into a classification problem of supervised learning. A considerable number of supervised learning algorithms for classification problems can be found in references. For our experimental validation, we selected four algorithms, i.e., decision tree inductions, back-propagation neural networks, support vector machines, and stochastic configuration networks.

In summary, for a specific problem of vibration signal diagnosis, there are two key factors, i.e., feature extraction and classification algorithm. Given each factor, one can find from references a considerable number of methodologies to complete the corresponding task. For example, Kumar et al. ([24]) illustrated that the performance of the feature selection method relies on the performance of the learning method. Tasmin et al. ([25]) used the five feature sets (Four feature sets through feature engineering and one original feature set), and they presented that for human activity recognition systems, data preprocessing and feature selection dramatically affect the classification performance in their paper. However, the four feature sets are datasets based on different feature selection approaches according to the same original dataset. In essence, it is the dimensionality reduction for the original features. In our study, we used different feature extraction methods rather than feature selection methods. Kitanovski et al. ([26]) used five different feature descriptors and three classification algorithms in the experiments. However, their experiments are to illustrate that not all visual features can be used to describe the relevant image patch correctly, and the best classification accuracy is achieved in the case of the GLDM descriptor. As pointed out in the paper title, we try to explore which one of the two key factors is more essential to achieve a better performance of learning and predicting. The exploration is basically experimental.

The rest of this article is organized as follows. Section 2 introduces an approach to extracting statistical features. Section 3 gives a method of generating parametric features based on Gaussian models. Section 4 summarizes several common algorithms of classification. Section 5 gives the method of stacking features. Section 6 conducts our experiments on a specified dataset and lists the corresponding consequences. Finally, section 7 provides the conclusion of this work.

## 2    Statistical features extraction from vibration signal

Vibration signal collected from equipment is usually considered to contain important information about the health status of the equipment. Preprocessing the original vibration and then extracting the statistical feature has been an indispensable step which is usually followed by a classification model in machine learning to complete a task of fault diagnosis [27,28,29]. Initially, statistical feature extraction from the vibration signal is actually a preprocessing process that, according to reference [30], can be specifically divided into three steps, as shown in Fig.1.

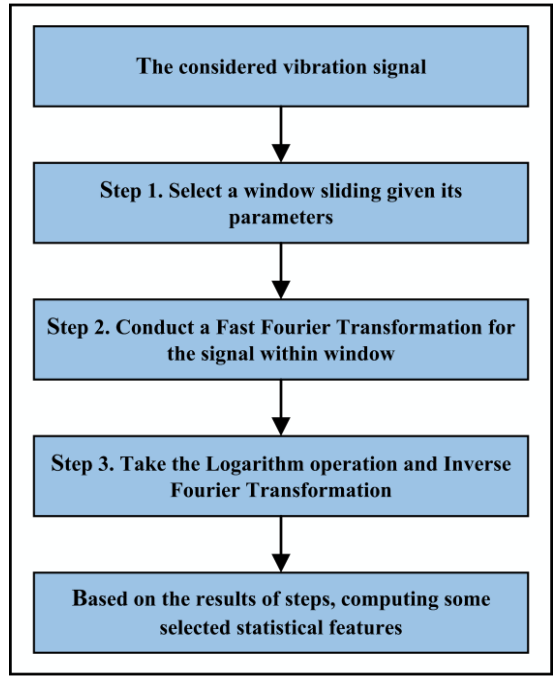

**Fig. 1.** Preprocessing procedures.

The first step is to design a window and use the window to divide the vibration signal. When processing time-domain signal, it is expected to extract features from a small window of the vibration signal. It usually has a rough assumption that the signal is stationary. The small window used in our study is the Hamming window, which shrinks the values of the signal toward zero at the window boundaries, avoiding discontinuities and the problem of spectrum leakage. In our study, the window parameters, i.e., the size of the window and the size of each shift, are specified as 250 and 100 data points, respectively. The equation of Hamming window is given as:

$$hamming\ w(t) = \begin{cases} 0.54 - 0.46\cos(\frac{2\pi t}{k}), & 0 < t < k-1 \\ 0, & otherwise \end{cases} \quad (1)$$

where the parameter values, 0.54 and 0.46, are empirical.

The second step is to conduct a Fast Fourier transform for the signal within the window. With the processing of the time domain signal with a sliding window, the time-domain signal with fixed frame length can be obtained. Fast Fourier transform (FFT) can help to know how much energy the signal contains at different frequency bands. It extracts information for frequency bands to form discrete-time values.

The third step is to take the Logarithm operation and Inverse Fourier transform. After FFT for the selected piece of signal, to perform the Inverse Fourier transform (IFFT) based on the logarithm(log) function, which can make the feature estimates less sensitive to variations in input. The waveform transformation of a frame signal is shown in Fig.2.

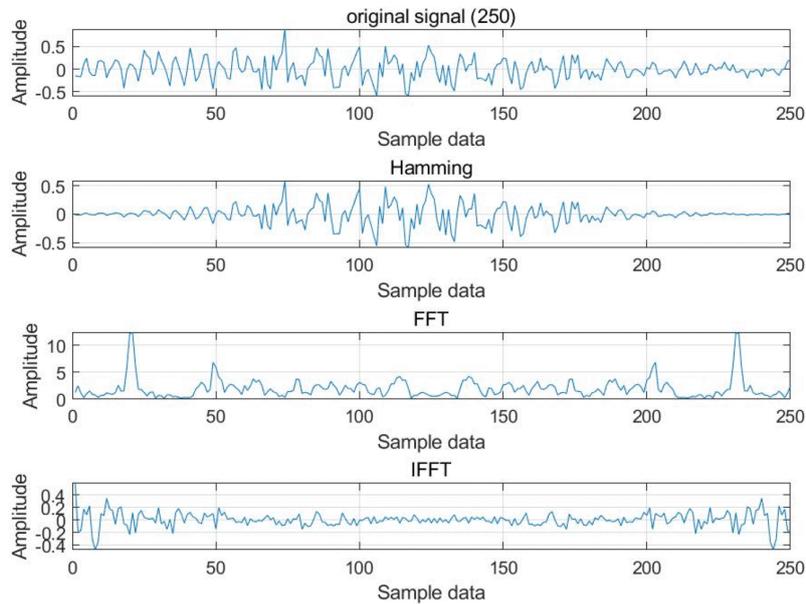

**Fig. 2.** The waveform transformation of a frame signal.

Based on the preprocessing results in the three steps mentioned above, we can extract some statistical features, and then the waveform signal is converted into a numerical vector. It is still empirical how many and what statistical features will be used in this process, which we should pay particular attention to.

In our study, we extract ten statistical features of the waveform. The specific information regarding the ten statistical features is shown in Table 1. In this way, each frame waveform is finally converted into a 10-dimension feature vector.

**Table 1.** Formula of ten statistical features.

| Feature | Formula | Feature | Formula |
|---|---|---|---|
| Mean | $x_\mu = \frac{1}{k}\sum_i^k x_i$ | Root mean squared | $x_{rms} = (\frac{1}{n}\sum_i^k x_i)^2$ |
| Standard deviation | $x_{std} = (\frac{1}{k-1}\sum_i^k (x_i - x_\mu)^2)^{\frac{1}{2}}$ | Crest factor | $x_{CF} = \frac{x_p}{x_{rms}}$ |
| Skewness | $x_{ske} = \frac{\sum_i^k (x_i - x_\mu)^3}{(n-1)x_{std}^3}$ | Shape factor | $x_{SF} = \frac{x_{rms}}{x_\mu}$ |
| Kurtosis | $x_{kt} = \frac{\sum_i^k (x_i - x_\mu)^4}{(n-1)x_{std}^4}$ | Impulse factor | $x_{IF} = \frac{x_p}{x_\mu}$ |
| Peak-to-peak | $x_p = \max(x) - \min(x)$ | Energy factor | $x_{EF} = \sum_i^k x_i^2$ |

## 3  Generating parametric features based on Gaussian models

Parametric features extraction based on Gaussian models is a fitting process of the signal waveform (e.g., Lu et al. [17] using a three-term Gaussian model to fit the pulse cycle). The fitting process can also be regarded as regenerating data from the multiple Gaussian distributions. The model of the $n$-term Gaussian function is given as:

$$g(x \mid A_1, \mu_1, \sigma_1, A_2, \mu_2, \sigma_2, \ldots, A_n, \mu_n, \sigma_n) = \sum_{i=1}^n A_i e^{-\left(\frac{x-\mu_i}{\sigma_i}\right)^2} + \varepsilon \qquad (2)$$

In Eq. (2), $n$ is the number of terms in the Gaussian models. $A_i$, $\mu_i$, $\sigma_i$ is the $i$-th function parameter, representing the amplitude, phase, and variance, respectively. $\varepsilon$ is the fitting error.

Gaussian models finish the fitting process of the signal waveform by adjusting the model parameters. The parametric features are usually using the model parameters. The specific method is shown in Fig.3.

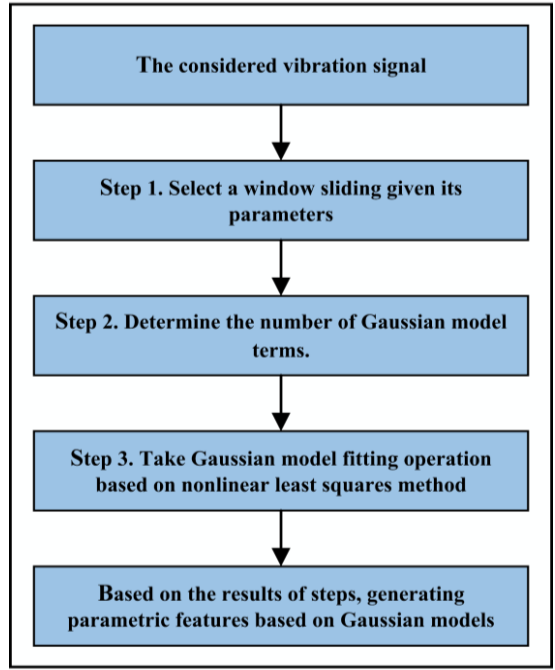

**Fig. 3.** Curve fitting based on Gaussian models.

The first step is to design a window and use the window to divide the signal. The window used in this step is the Hamming window, which is the same as Section 2. The window parameters, i.e., the size of the window and the size of each shift, are specified as 250 and 100 data points, respectively.

The second step is to determine the number of Gaussian model terms. Theoretically, it usually has a rough assumption that the Gaussian model with enough terms and the proper weight can fit arbitrary data distribution. Our study uses several kinds of Gaussian models to fit the vibration signal waveform, respectively.

The third step is using the nonlinear least-squares method for Gaussian curve fitting. The nonlinear least-squares method evenly calculates the difference between the actual value of the dependent variable and the value predicted by the model, i.e., the residuals. Lu et al. [17] mainly discussed the three-term Gaussian function with weights, but the weight is not a contribution in our study. The least-squares sense is expressed as:

$$min(x)\|f(x_i) - y_i\|^2, i = 1, 2, \ldots, n \quad (3)$$

In Eq.2. $f(x_i)$ is the Gaussian function value in the $i$-th data point, and $y_i$ is the actual value in $i$-th data point $x_i$. $n$ represents the number of data points.

Fig.4. shows the fitting results of Gaussian models. Due to the limitation of the number of terms in Gaussian models, the result is not exactly the same as the signal

waveform. It can be seen that the peak density of the signal waveform is different in different places. The Gaussian model prefers to fit the whole waveform and overall describe the changes in different signal regions.

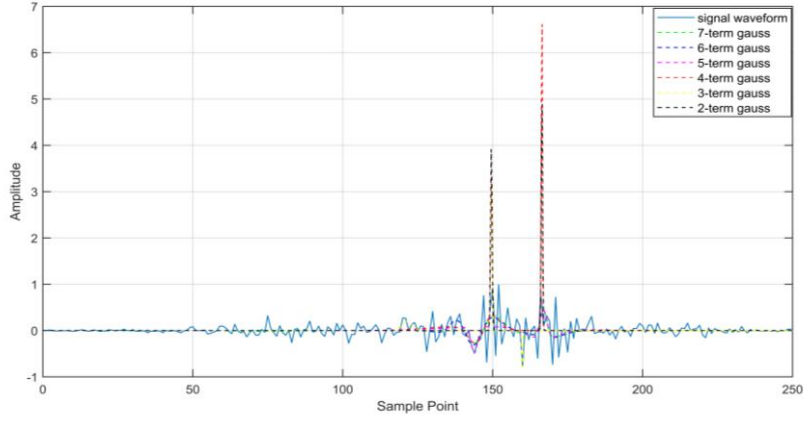

**Fig. 4.** Comparison of different-term gaussian function.

Finally, we can obtain the parametric features of Gaussian models based on the fitting results. In Eq. (2), each Gaussian function has three parameters, i.e., $A, \mu, \sigma$. Obviously, the number of extracted features depends on $n$-term Gaussian functions. For the n-term term Gaussian, there are $3n$ features to be extracted. The specific parametric features are shown in Table 2.

**Table 2.** Parametric features based on Gaussian model.

| n-term Gaussian | Parametric features |
| --- | --- |
| 7-term | $A_1, \mu_1, \sigma_1, \ldots, A_7, \mu_7, \sigma_7$ |
| 6-term | $A_1, \mu_1, \sigma_1, \ldots, A_6, \mu_6, \sigma_6$ |
| 5-term | $A_1, \mu_1, \sigma_1, \ldots, A_5, \mu_5, \sigma_5$ |
| 4-term | $A_1, \mu_1, \sigma_1, \ldots, A_4, \mu_4, \sigma_4$ |
| 3-term | $A_1, \mu_1, \sigma_1, \ldots, A_3, \mu_3, \sigma_3$ |
| 2-term | $A_1, \mu_1, \sigma_1, A_2, \mu_2, \sigma_2$ |

## 4 Several classification algorithms

This section briefly introduces several classification algorithms in supervised learning. More details can be found from references [31,33,35,38]. There selected classification algorithms will be used to validate the performance of extracted features.

## 4.1 Decision Tree

Top-down induction of decision trees [31] is a type of inductive learning algorithm, which aims to generate a set of rules represented in a tree structure and has been applied in various fields such as fault diagnosis [32]. The basic idea of learning a decision tree is to recursively select an attribute on which each internal node of the tree is split, until a leaf node is obtained due to the satisfaction of the stopping criteria (to be specified below). Some popularly used heuristics of attribute selection include conditional entropy (Eq. (4)) and Gini-index (Eq. (5)).

$$CE = -\sum_i p(x=v_i) \sum_c p(y=c \mid x=v_i) \log_2 p(y=c \mid x=v_i) \qquad (4)$$

$$GI = \min_i p(x=v_i)\left(1 - \sum_c p(y=c \mid x=v_i)^2\right) + p(x \neq v_i)\left(1 - \sum_c p(y=c \mid x \neq v_i)^2\right) \qquad (5)$$

In Eq. (4) and Eq. (5), $p(x = v_i)$ represents the probability that the value of a symbolic feature x is $v_i$ and $p(y = c|x = v_i)$ denotes the posterior probability of an instance belonging to class c given the condition that $x = v_i$.

No matter which one of the above-mentioned heuristics is used for attribute selection, the criteria of stopping the growth of a decision tree would be satisfied as soon as a node covers only one class of training instances or all the attributes have previously been selected for those parent nodes of the current node.

## 4.2 Neural Network

The Neural network is a type of computational model that can be learned through the backpropagation (BP) algorithm [33] and has been used in various application areas such as fault diagnosis [34]. A neural network model typically consists of an input layer, an output layer, and at least one hidden layer, where each layer has one or more nodes, and each node in layer $l_j$ is connected to some or all nodes in the preceding layer $l_{j-1}$. The basic idea of learning a neural network model is to optimize the weights (parameters) of the connections between nodes in any two adjacent layers until the loss calculated in the output layer has been minimized. The optimization of the parameters is typically achieved by using the stochastic gradient descent method (or its variants), as illustrated in Eq. (6), Eq. (7), and Eq. (8).

$$h_{j,k} = \delta\left(\sum_{k'} w^t_{k,k'} \cdot h_{j-1,k'}\right) \qquad (6)$$

$$loss = (y - h_{-1})^2 \qquad (7)$$

$$w^{t+1}_{k,k'} = w^t_{k,k'} - \eta \frac{\partial loss}{\partial w^t_{k,k'}} \qquad (8)$$

In Eq. (6), $h_{j,k}$ denotes the output of the $k$-th node in the $j$-th layer, and $w^t_{k,k'}$ represents the weight updated at iteration t for the connection between the $k$-th node in the $j$-th layer and the $k^{\wedge\prime}$-th node in the $(j-1)$-th layer, and $h_{j-1,k'}$ denotes the output of the $k'$-th node in the $(j-1)$-th layer. Eq. (7) represents a mean square error measured by computing the difference between the true output y and the predicted output $h_{-1}$ obtained at the only node in the output layer in a regression context. Eq. (8) represents how the parameter $w_{k,k'}^{t+1}$ is updated at a new iteration t+1, where $\eta$ is the learning rate set as a tuning parameter.

### 4.3 Support Vector Machine

Support vector machine (SVM) [35] is a type of kernel method, which aims to build a margin to separate two classes and has been applied in various fields such as fault diagnosis [36]. The basic idea of learning an SVM model is to find a maximum margin hyperplane to minimize the risk of incorrect classification, i.e., it is a constrained convex optimization problem as illustrated in Eq. (9). Moreover, the original representation of data may result in a linearly non-separable feature space. In this case, a kernel function can be used to map a linearly non-separable feature space into a higher dimensional space that is linearly separable, as illustrated in Eq. (10).

$$\min_{w,b} = \frac{1}{2}\|w\|^2 \; s.t. \; y_i(wx_i + b) \geq 1, i = 1, 2, \ldots, N \tag{9}$$

$$K(x_i, x_j) = \varphi(x_i) \cdot \varphi(x_j) \tag{10}$$

In Eq. (9), w and b are two parameters that need to be optimized towards obtaining a maximum margin hyperplane, where $x_i$ and $y_i$ represent the feature vector and the ground truth label of the $i$-th instance, respectively. Eq. (10) represents that a mapping $\varphi$ exists such that the output of a kernel function $K(x_i, x_j)$ is equal to the inner product of the two transformed feature vectors $\varphi(x_i)$ and $\varphi(x_j)$ for any two instances $x_i$ and $x_j$.

### 4.4 Stochastic Configuration Network

Stochastic configuration network (SCN) [37] is a type of network model based on the randomized algorithm, which has been used in various application areas such as fault diagnosis [38]. The basic idea of SCN is to set randomly all the weights for the connections between nodes in the hidden layers and not to update these weights in the whole training process. The only thing that needs to be optimized in the learning process is the set of output weights for the connections to the node(s) in the output layer, where the output function $f$ is illustrated in Eq. (11) and Eq. (12).

$$f = \sum_m \beta_m \varphi_m(x) = \varphi(x)\beta \tag{11}$$

$$\varphi(x) = G(w_m, b_m, x) \tag{12}$$

In Eq. (11), $\beta_m$ is the output weight from the $m$-th node in the last hidden layer, and $\phi_m$ is a function that maps the input feature value x to a new feature value $\phi_m(x)$ as the output of the $m$-th node in the last hidden layer. Eq. (12) shows the specific design of the feature mapping function, where $w_m$ denotes the set of input weights and $b_m$ is the bias term for the $m$-th node in the last hidden layer.

## 5    Feature stacking

In order to compare the performances of statistical features and parametric features, we have conducted a great deal of vibration signal diagnosis pre-experiments based on SCN under various working conditions. The results are shown in Table.3.

**Table 3.** The testing accuracy of pre-experiments base on SCN.

| Working conditions | | Statistical feature | 7-term Gauss | 6-term Gauss | 5-term Gauss | 4-term Gauss | 3-term Gauss | 2-term Gauss |
|---|---|---|---|---|---|---|---|---|
| 10Hz | Fs5120 | **99.96%** | **77.38%** | **86.03%** | **82.29%** | **68.18%** | **78.40%** | **63.38%** |
| | Fs10240 | 99.98% | 92.24% | 98.21% | 85.30% | 68.62% | 67.68% | 55.05% |
| | Fs12800 | 99.87% | 95.02% | 76.88% | 52.83% | 49.71% | 49.81% | 51.88% |
| | Fs20480 | 99.97% | 96.80% | 79.41% | 85.34% | 86.39% | 70.01% | 55.31% |
| | Fs25600 | 99.76% | 99.79% | 99.73% | 98.50% | 92.42% | 76.25% | 77.78% |
| | Fs51200 | 99.97% | 99.61% | 99.12% | 99.62% | 97.92% | 98.95% | 64.61% |
| 20Hz | Fs5120 | 99.94% | 97.99% | 89.24% | 96.27% | 96.08% | 79.04% | 72.81% |
| | Fs10240 | **96.97%** | **71.14%** | **70.61%** | **69.43%** | **72.89%** | **91.34%** | **66.57%** |
| | Fs12800 | **88.83%** | **91.54%** | 88.24% | 86.65% | 83.13% | 68.00% | 90.18% |
| | Fs20480 | **89.83%** | **98.55%** | **96.42%** | **97.06%** | **98.91%** | 53.04% | 81.82% |
| | Fs25600 | 95.46% | 97.04% | 99.65% | 97.82% | 94.14% | 93.08% | 78.26% |
| | Fs51200 | 99.99% | 99.64% | 98.13% | 94.23% | 89.40% | 95.01% | 64.50% |
| 30Hz | Fs5120 | **100.00%** | **97.21%** | **84.71%** | **92.12%** | **79.39%** | **82.18%** | **35.96%** |
| | Fs10240 | **95.84%** | **87.26%** | **90.67%** | **96.28%** | **88.26%** | **68.77%** | **68.71%** |
| | Fs12800 | 82.96% | 97.11% | 96.56% | 96.95% | 87.70% | 92.88% | 79.92% |
| | Fs20480 | 84.76% | 99.64% | 99.18% | 98.80% | 82.27% | 95.46% | 83.38% |
| | Fs25600 | 96.54% | 99.58% | 97.92% | 96.07% | 90.29% | 86.04% | 75.77% |
| | Fs51200 | 99.90% | 98.38% | 96.09% | 98.55% | 92.39% | 93.21% | 60.15% |
| 40Hz | Fs5120 | 99.92% | 91.73% | 69.61% | 58.43% | 54.69% | 57.71% | 45.96% |
| | Fs10240 | 89.39% | 97.74% | 79.32% | 92.67% | 92.58% | 90.00% | 60.44% |
| | Fs12800 | **90.85%** | **86.66%** | 93.69% | 98.50% | **83.20%** | **85.07%** | **41.15%** |
| | Fs20480 | **86.82%** | **99.83%** | **99.76%** | 98.94% | **99.54%** | **94.21%** | 69.03% |
| | Fs25600 | **80.33%** | **99.66%** | **99.53%** | **98.62%** | **98.90%** | **91.17%** | 63.46% |
| | Fs51200 | 99.50% | 99.39% | 98.94% | 96.92% | 96.33% | 94.49% | 71.21% |
| 50Hz | Fs5120 | 99.98% | 82.87% | 80.57% | 72.39% | 82.63% | 81.26% | 52.57% |
| | Fs10240 | 99.34% | 91.95% | 91.32% | 94.60% | 73.18% | 63.13% | 62.51% |
| | Fs12800 | 98.11% | 91.11% | 98.98% | 98.88% | 90.66% | 99.09% | 77.86% |
| | Fs20480 | **83.72%** | **99.17%** | **99.53%** | **98.41%** | **94.57%** | **85.25%** | **83.72%** |
| | Fs25600 | 92.49% | 99.64% | 99.53% | 99.35% | 98.34% | 93.03% | 73.02% |
| | Fs51200 | 94.15% | 98.94% | 99.53% | 99.14% | 91.54% | 92.18% | 94.31% |

From the Table.3, there is a phenomenon worth considering here: the performance of the two types of features may roughly present a complementary situation under some working conditions. For instance, the accuracy of statistical features is 80.33% in the condition of the 40Hz-Fs25600, and the accuracy of the parametric features seems better than it. However, the accuracy of statistical features is better than the parametric features in the 20Hz-Fs10240 condition. Some typical samples have been in bold type. In our study, we further use a simple strategy, i.e., feature stacking. We stack the statistical features and the Gaussian parametric features. The specific feature stacking process is shown in Fig.5.

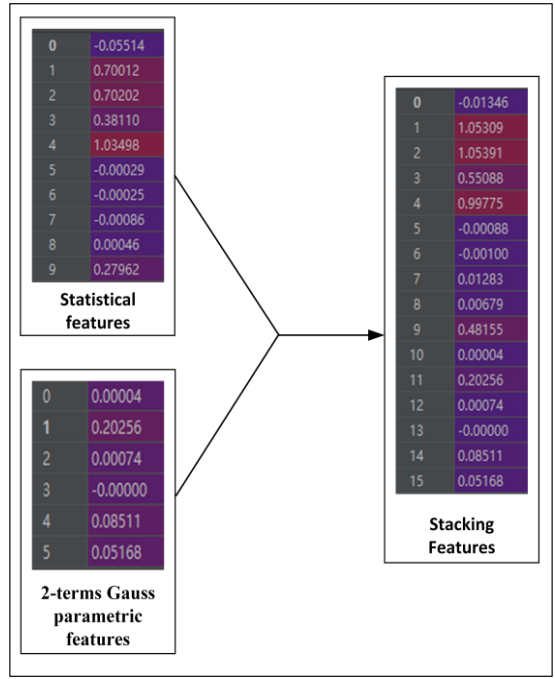

**Fig. 5.** Feature stacking.

In Fig.5, it is important to note that we should stack the two types of features from the same window of the vibration signal. The statistical feature in Section 2 is a 10-dimension feature vector, and the parametric feature based on the 2-term Gaussian model is 6-dimension, so the stacking feature in Fig.5 is the 16-dimension feature vector. The same window is used in the process of feature extraction, and the measure ensures the efficiency of feature stacking. In addition, the maximum dimension of the stacking features is 31, which is reasonable for the size of the window. There are different dimensions of the final generating stacking features. The dimensions of the stacking features are shown in Table 4.

Table 4. The dimensions of stacking features.

| Features | Dimensions |
|---|---|
| Statistical features + 7-terms Gauss | 31 |
| Statistical features + 6-terms Gauss | 28 |
| Statistical features + 5-terms Gauss | 25 |
| Statistical features + 4-terms Gauss | 22 |
| Statistical features + 3-terms Gauss | 19 |
| Statistical features + 2-terms Gauss | 16 |

## 6   Experimental analysis

### 6.1   Introduction to used data set

The data used in this study is generated based on the Drivetrain Diagnostics Simulator (DDS), which is manufactured by Spectra Quest company in America. The data acquisition equipment mainly consists of both hardware and software systems. The hardware is the Dong-fang data acquisition instrument which is 51.2kHz of the maximum sample frequency. The sensor is Piezoelectric IEPE acceleration. The software system adopts DASP-V11 (Engineering version) software platform.

In our study, the experimental object is a planetary gearbox, and the bearing type is Rexnord ER16K. The transmission ratio is 4.5714, and the transmission ratios of the first stage spur gear and the second stage spur gear are 3.4483 and 2.5, respectively. The experimental conditions are set as follows: rotating speed is 10-20-30-40-50 Hz, and sampling frequency (FS) is 5120-10240-12800-20480-25600-51200 Hz. The class consists of the normal state and four fault types ('Outer race', 'Inner race', 'Ball', 'Combo'). The sampling time for every condition is 10 seconds. The original vibration signal was stored in the text, and the file name marks the information (the type of fault, the rotating speed, the sampling frequency). The text files are shown in Fig.6. In the text, the first column is sampling time, and the second and the third are the original vibration signal from two different sensors, respectively. So each text can obtain the datasets of two sensors.

[File listing image showing signal data files organized by fault type (Ball, ComBO, Inner, Normal, Outer) across different Hz values (10-50Hz) and sampling frequencies (Fs5120 to Fs51200)]

**Fig. 6.** Signal data file.

Finally, the features can be extracted from vibration signal data. The statistical features are extracted in Section 2, and we generate parametric features with several different kinds of dimensions based on Gaussian models in Section 3. Then, the stacking feature with several different kinds of dimensions can be obtained in section 5. So, thirteen datasets can be obtained. The number of samples in each dataset is no different. The 50Hz_Fs20480 condition is used as an example to illustrate our experimental results. The specific number of each class sample is shown in Table 5.

**Table 5.** The data sets used in the experiment.

| Category | Total |
|---|---|
| Outer fault | 1683 |
| Inner fault | 1683 |
| Combo fault | 1708 |
| Ball fault | 1708 |
| Normal | 1683 |

### 6.2 Experimental task

In the experiments, we use the data sets shown in Section 6.1. We define the tasks as a multi-classification problem, and there are 5-class in our experiments of vibration signal diagnosis. The training and test sets for each class fault are divided in the proportion 4:1, and cross-validation is used.

Our pre-experiments were mainly carried out in SCN. We used the grid search method to determine the optimal number of hidden layer nodes under different working conditions, and the grid search range is 10-100. We find that the number of nodes varies significantly under different working conditions. The minimum is only 12, and the maximum is 100. Considering the training cost, we have chosen 40 as the final number of hidden nodes. In subsequent experiments, we found that the SCN network with 40 hidden nodes performs well. Besides, we set the other parameters based on the algorithm in reference [33] given by Wang et al.

At the same time, we also use 40 as the number of hidden layer nodes in the BP neural network for comparison. The number of classes determines the number of output layer nodes, and the number of input layer nodes is set according to the dimensions of different features. The fixed learning rate is 0.0075 to ensure convergence and good performance. In our experiments, the results of statistical features are not good in the BP neural network, so we have to make many iterations, and the final number of iterations is 200000.

It is crucial to select the kernel function of SVM. We use the radial basis function (RBF) as kernel to ensure running speed and reduce cost. In order to maintain the balance between overfitting and underfitting, the kernel function parameter $\gamma$ is 0.1, and C is 0.8.

For the decision tree, some popularly used heuristics of attribute selection include entropy and Gini-index, and the choice is a little different in general. Entropy is used in our study, and the other parameters of the classifier are default values.

Most parameters in our experiments are empirical. Although these parameters may not be optimal, they had achieved good performance in our study.

### 6.3  Experimental result

Table 6 shows the testing accuracy of different features under the four classifiers for vibration signal faults. We can see the prediction accuracies.

Firstly, we illustrate the results from the perspective of classifiers. For the decision tree, the classification accuracy of each kind of feature in the decision tree shows high performance, and the accuracy of statistical features is relatively low. For BP neural network, the classification accuracy of statistical features and 2-term Gaussian parametric features is 83.19% and 88.67%, respectively, where others perform better. For SVM, the classification accuracy of statistical features is 48.44% which is a terrible result, and the 3-term and 2-term Gaussian parametric features are low than others. For SCN, the performance of statistical features, 2-term, and 3-term Gaussian parametric features are worse than others.

Secondly, we illustrate them from the perspective of features. Statistical features did not perform well in different classifiers, and the worst result is 48.44% in SVM. For parametric features based on the Gaussian model, the performance of 3-term and 2-term Gaussian parametric features may be less potent than others, but they can be acceptable. Moreover, the classification accuracy showed no apparent differences among 7-term, 6-term, and 5-term parametric features, and the 6-term Gaussian par-ametric features might perform a little better than 7-term. In addition, all of the stacking features show higher performance for stacking features.

Finally, we can see that the stacking features can perform better. For example, the classification accuracy of statistical features is 48.44% in SVM, and the accuracy of 3-term Gaussian parametric features is 84.25%, but they generate the stacking features is 91.92% of the classification accuracy. For the stacking features based on the 7-term and 6-term Gaussian parametric features, the accuracies are better than or similar to those of parametric features. In addition, the performance of stacking features is more stable under the different classification algorithms.

**Table 6.** Accuracy of classifiers according to testing sets with different features.

| Features | | Decision Tree | BP-Neural network | SVM | SCN |
|---|---|---|---|---|---|
| Statistical features | | 90.79% | 83.19% | 48.44% | 83.72% |
| 7-term Gauss | parametric features | 99.88% | 99.71% | 99.82% | 99.35% |
| | stacking features | **99.94%** | **99.76%** | 99.70% | 99.17% |
| 6-term Gauss | parametric features | 99.76% | 99.94% | 99.88% | 99.59% |
| | stacking features | 99.76% | 99.94% | 99.88% | **99.65%** |
| 5-term Gauss | parametric features | 99.53% | 99.94% | 99.17% | 98.41% |
| | stacking features | **99.59%** | 99.94% | **99.23%** | **99.53%** |
| 4-term Gauss | parametric features | 98.41% | 96.52% | 94.93% | 94.57% |
| | stacking features | **98.64%** | **99.88%** | **97.46%** | **99.47%** |
| 3-term Gauss | parametric features | 98.05% | 98.29% | 84.25% | 84.25% |
| | stacking features | **98.34%** | **98.58%** | **91.92%** | **98.29%** |
| 2-term Gauss | parametric features | 98.35% | 88.67% | 83.19% | 83.72% |
| | stacking features | **98.58%** | **99.23%** | **92.68%** | **99.35%** |

### 6.4 Analysis and discussion

In Table 6, we can know the contribution of stacking features intuitively. The stacking features improve the classification accuracy in the classifier. In the same classifier, the stacking features can obtain higher accuracy. For the different classifiers, the stacking features perform more stable.

In Fig.4, it can be found that the fitting effect on the vibration signal waveform is not as well as possible. However, the parametric features based on Gaussian models can still obtain relatively high accuracy in the classifiers. Maybe it is because the Gaussian models tend to describe the whole segments of the signal waveform.

The classification accuracy of the 6-term Gaussian features in the classifiers might perform a little better than others. Maybe it can also be considered that the performance of Gaussian features depends on the number of the terms to some extent. Generally, this does not mean that better results can always be achieved with more ex-tracted features because the 7-term Gaussian features did not perform better than the 6-terms that may be due to overfitting. Moreover, the classification accuracy of stacking features based on 7-term Gaussian features might not all be higher than others. For instance, the stacking features based on the 4-term Gaussian features perform better than the stacking features based on 7-term Gaussian features in SCN. Indeed, it is not easy to illustrate that the more features can get the best result. The dimensions of the stacking features based on 7-term or 4-term Gaussian parametric features are 31 and 22, respectively, but it cannot be able to testify that better results can always be achieved with more extracted features. In addition, it is not difficult to know that the more terms of the Gaussian model used, the more cost will be brought. The stacking features get rid of the dependence of Gaussian features on the number of terms to a certain extent and decrease cost.

The stacking features compared with the statistical features or the parametric features based on the Gaussian model can obtain high classification accuracy without significant fluctuation in the four classifiers. It can be considered as the stacking features reduce the requirements of the classification algorithm in the task of vibration signal fault diagnosis.

## 7  Conclusions

In this paper, we use the statistical features, parametric features based on Gaussian models, and stacking features to accomplish the experimental study of vibration signal diagnosis. The experiment results show that an effective feature extraction method has a greater impact on the classification accuracy compared with the classification algorithm, and more effective features can reduce the dependence on classification algorithms to a certain extent. Specifically, if the extracted features are good enough, any classification algorithm may perform very well. Unfortunately, it is a really challenging task that is almost impossible to fulfill in the real application.

In this paper, the initially extracted features can be regarded as two different modes which are stacked together, resulting in a perfect prediction almost independent of classification algorithms. In deep learning mechanisms for practically complicated problems, these initial features are usually put into a multilayer neural network for transformations in order to achieve higher level features.

Acknowledgement

This work was supported in part by the National Natural Science Foundation of China under Grant 51807124, in part by the Youth Talent Project of China's Hebei Provincial Education Department under Grant BJ2020054 and the Youth Foundation of Hebei science and technology research project QN2018108.